\documentclass[lettersize,journal]{IEEEtran}
\usepackage{amsmath,amsfonts}
\usepackage{algorithmic}
\usepackage{array}
\usepackage[caption=false,font=normalsize,labelfont=sf,textfont=sf]{subfig}
\usepackage{textcomp}
\usepackage{stfloats}
\usepackage{url}
\usepackage{multirow}
\usepackage{multicol}
\usepackage{verbatim}
\usepackage{graphicx}
\usepackage{booktabs}
\usepackage{color}
\usepackage{threeparttable}

\def\BibTeX{{\rm B\kern-.05em{\sc i\kern-.025em b}\kern-.08em
    T\kern-.1667em\lower.7ex\hbox{E}\kern-.125emX}}
\usepackage{balance}

\begin{document}

\title{Leveraging BEV Representation for\\360-degree Visual Place Recognition}
\author{Xuecheng Xu$^{1,3}$, Yanmei Jiao$^{2}$, Sha Lu$^{1}$, Xiaqing Ding$^{3}$, Rong Xiong$^{1}$, and Yue Wang$^{1}$ 
\thanks{$^{1}$Xuecheng Xu, Sha Lu, Rong Xiong and Yue Wang are with the State Key Laboratory of Industrial Control Technology and Institute of Cyber-Systems and Control, Zhejiang University, Hangzhou, 310058, China. Yue Wang is the corresponding author {\tt\small wangyue@iipc.zju.edu.cn}.}
\thanks{$^{2}$Yanmei Jiao is with the School of Information Science and Engineering, Hangzhou Normal University, Hangzhou, 311121, China.}
\thanks{$^{3}$Xiaqing Ding is with Alibaba Group, Hangzhou, 310052, China. This work is done when Xuecheng Xu is an intern at Alibaba Group.}}

\markboth{Journal of \LaTeX\ Class Files,~Vol.~18, No.~9, May~2023}%
{How to Use the IEEEtran \LaTeX \ Templates}

\maketitle

\begin{abstract}
This paper investigates the advantages of using Bird's Eye View (BEV) representation in 360-degree visual place recognition (VPR). We propose a novel network architecture that utilizes the BEV representation in feature extraction, feature aggregation, and vision-LiDAR fusion, which bridges visual cues and spatial awareness. Our method extracts image features using standard convolutional networks and combines the features according to pre-defined 3D grid spatial points. To alleviate the mechanical and time misalignments between cameras, we further introduce deformable attention to learn the compensation. Upon the BEV feature representation, we then employ the polar transform and the Discrete Fourier transform for aggregation, which is shown to be rotation-invariant. In addition, the image and point cloud cues can be easily stated in the same coordinates, which benefits sensor fusion for place recognition. The proposed BEV-based method is evaluated in ablation and comparative studies on two datasets, including on-the-road and off-the-road scenarios. The experimental results verify the hypothesis that BEV can benefit VPR by its superior performance compared to baseline methods. To the best of our knowledge, this is the first trial of employing BEV representation in this task.

\end{abstract}

\begin{figure}[tbp]
	\centering
		\includegraphics[width=\linewidth]{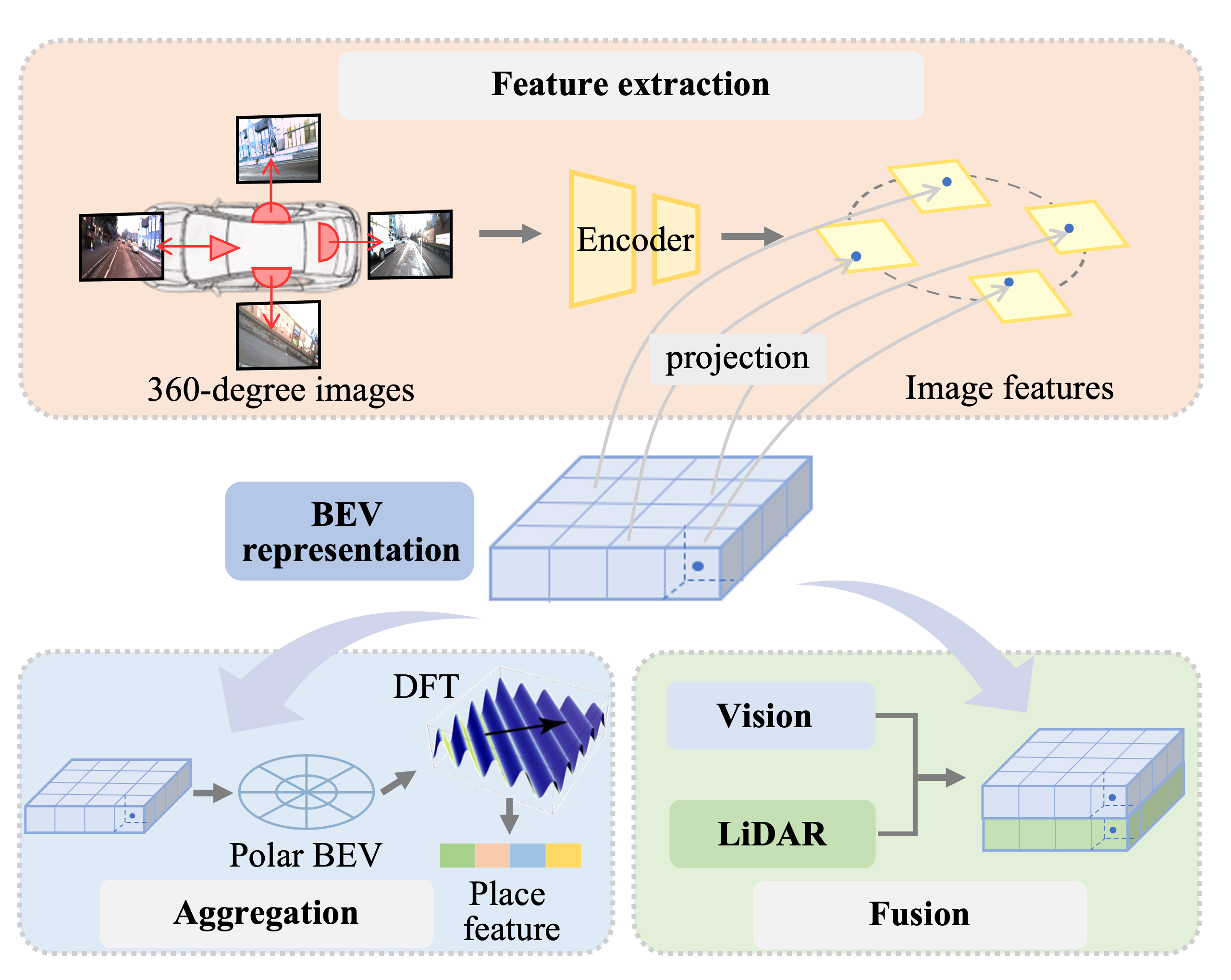}
	\caption{The BEV representation for 360-degree visual place recognition is introduced. The BEV representation benefits the VPR in three ways. First, in the feature extraction part, a standard convolutional network is better suited to multi-view images than panoramic images. Second, rotation invariance on the BEV representation can be easily achieved by the combination of the polar transform and the Discrete Fourier transform. Third, it is more plausible to fuse the image and point cloud features within the same coordinate system.}
	\label{fig:Teaser}
    \vspace{-0.4cm}
\end{figure}

\section{Introduction}
Visual place recognition (VPR) is an essential component of global localization in the field of autonomous driving. The VPR task is challenging because it must deal with the entangled variation from environmental changes such as illumination and seasonal transitions, as well as large perspective differences. Thus, learning a robust place representation becomes a long-standing problem in the community \cite{arandjelovic2016netvlad, doan2019scalable, torii201524, chen2017deep, chen2018learning, garg2022semantic, khaliq2019holistic, lowry2015visual}.

In the VPR literature, both classification \cite{chen2017deep, cao2020unifying} and contrastive representation learning \cite{arandjelovic2016netvlad, revaud2019learning} algorithms have been used to generate a single image-based place feature for efficient retrieval. Some of these methods have already been employed in autonomous systems that operate in well-conditioned situations \cite{lajoie2020door, ebadi2022present}. 
As the images taken from a single perspective camera have narrow fields of view, such a solution is incapable of covering the surrounding scene. Therefore, place recognition only happens if the place is revisited from a viewpoint that is quite similar to the original one. 
Fortunately, with the increasing expansion of autonomous driving, multi-camera setups are gaining popularity and becoming more affordable \cite{caesar2020nuscenes, barnes2020oxford, sun2020scalability}. Multiple cameras deployed on autonomous vehicles are able to capture the nearly 360-degree view, equivalent to LiDAR, thus promising the robustness of VPR when a large perspective occurs.

The most straightforward way to tackle the VPR task in a multi-camera setup is to generate panoramic images from multi-camera images and employ the existing VPR methods, which mainly aggregate features from the image plane. Alternatively, we note that the bird's-eye view (BEV) representation attracts recent attention. Several pioneering works \cite{li2022bevformer,liu2022bevfusion} demonstrate that features from multi-camera images can be aggregated in BEV representation for better performance in detection and segmentation compared with image-based methods. Such a trend raises a question: \textit{Can BEV representation benefits visual place recognition?}



In this paper, we set to study the effective pathways that BEV representation can promote visual place recognition. We consider that the BEV representation has advantages in three aspects: feature extraction, feature aggregation, and the vision-LiDAR feature fusion since these parts can be endowed with spatial awareness. Specifically, as shown in Fig.~\ref{fig:Teaser}, we first extract features from images using standard convolutional networks. After the image feature extraction, pre-defined 3D spatial points are then projected onto the images to further extract the spatial features. To alleviate the possible misalignment brought by the noisy calibration or synchronization, we further employ deformable attention. Then the consolidated features in 3D space are then converted to polar BEV, where the Discrete Fourier transform is used to achieve rotation-invariant feature aggregation. Upon BEV representation, we state the image and point cloud cues in the same coordinate, which benefits the sensor fusion for place recognition. In the experiments, we evaluate the proposed BEV-based method in ablation and comparative studies. The results show that superior performance can be achieved using our methods, verifying the hypothesis. In summary, the contributions are three-folded:
\begin{itemize}
    \item We investigate the benefits of BEV representation in 360-degree visual place recognition. To the best of our knowledge, this is the first trial of employing such representation in this task.
    \item We propose a network architecture to utilize the BEV representation in feature extraction, feature aggregation, and vision-LiDAR fusion, bridging the visual cues and spatial awareness.
    \item We verify the hypothesis that BEV benefits VPR in two datasets including off-the-road and on-the-road scenarios. The method shows competitive performance compared with the baseline methods, demonstrating its effectiveness. Our code is available at https://github.com/MaverickPeter/vDiSCO.
\end{itemize}


\section{Related Works}
\subsection{Visual Place Recognition}

Early VPR methods often use local features such as SIFT \cite{valgren2007sift}, SURF \cite{bay2006surf} and ORB \cite{rublee2011orb} with aggregate strategies like Bag of Words (BoW) \cite{csurka2004visual}, Vector of Locally Aggregated Descriptors (VLAD) \cite{arandjelovic2013all}, or Aggregated Selective Match Kernels (ASMK) \cite{tolias2013aggregate}. These local features can be easily applied to panoramic images. \cite{chapoulie2011spherical} proposes visual loop closing methods that adopt SIFT as local features and BoW as aggregation. Such an implementation is an imitation of the common VPR pipeline, which also has issues with feature discrimination.
With the prevalence of deep learning, remarkable progress has been made in local features \cite{detone2018superpoint, dusmanu2019d2, balntas2016learning}. Most known local features are reviewed in \cite{zhou2017recent, chen2022deep}. In recent local feature learning frameworks, DELF \cite{noh2017large} applies an attention layer to select local keypoints, and DOLG \cite{yang2021dolg} further utilizes multi-atrous convolution layers to include multi-scale feature maps. 

Other than aggregated by local features, global descriptors can also be generated by several differentiable aggregation operations, such as sum-pooling \cite{tolias2020learning}, GeM pooling \cite{radenovic2018fine} and NetVLAD \cite{arandjelovic2016netvlad}. Thanks to the strong ability provided by the end-to-end CNN model, NetVLAD \cite{arandjelovic2016netvlad} and its variants \cite{yu2019spatial, hausler2021patch, khaliq2022multires} outperform early methods. Several methods \cite{orhan2021efficient, fang2020cfvl} adopt the CNN model to extract discriminative features on panoramic images. However, convolutional networks may not be the best option because objects in panoramic images tend to be distorted. Besides feature extraction and aggregation, many losses, including ranking-based triplet \cite{bronstein2010data}, quadruplet \cite{chen2017beyond}, and listwise losses \cite{revaud2019learning} are proposed in order to improve the training of neural networks. More recently, some approaches focus on re-ranking the place matches using sequential information \cite{milford2012seqslam, garg2021seqnet}, query expansion \cite{chum2011total} or geometric verification \cite{garg2022semantic,noh2017large}. With the fast development of LiDAR place recognition, researchers are also interested in multi-modal fusion methods. Multi-modal methods \cite{komorowski2021minkloc++,lai2021adafusion} bring LiDAR to vision and thus improve the performance in some degraded scenarios. These approaches, however, merely combine descriptors from different modalities and ignore the connections between their distinct features.
 
As mentioned above, the majority of previous works concentrate on feature extraction and aggregation. In contrast to these approaches, we concentrate on feature representation. Leveraging the BEV representation in feature extraction, feature aggregation, and vision-LiDAR fusion, we bring the spatial awareness to the visual cues.

\subsection{BEV Representation}

The majority of previous VPR techniques are designed to handle retrieval tasks with single-view images. With the advancement of autonomous robots, multi-view images are becoming more accessible. However, the corresponding VPR algorithms using multi-view images have not yet been developed. In contrast to single-camera setups, representation in multi-camera scenarios can be diverse. Panorama as an intuitive representation of multi-view images is only occasionally mentioned in the literature \cite{chapoulie2011spherical}, not to mention other representations. On the contrary, in the field of autonomous driving, various spatial representations have recently gained popularity due to the rapid growth of 360-degree vision suites. Representations such as BEV and 3D volume perform better than separate images because features in the image plane are transformed into features in a spatial representation, bringing spatial awareness across 360-degree images. Among all representations, BEV is a commonly used one since it clearly presents the location and scale of objects and is suitable for various autonomous driving tasks. In the related LiDAR place recognition, several methods \cite{kim2021scan, xu2021disco, chen2021overlapnet, xu2022ring++} utilize the BEV representation to achieve rotation invariance and estimation. In \cite{ding2022translation}, authors further prove that, in contrast to other representations like range images, the translation variance in BEV can be decoupled and eliminated. Some works also explore the possibility of leveraging multi-modal information using unified BEV representation. CORAL \cite{pan2021coral} creates a multi-modal BEV that works better than LiDAR-only techniques by projecting image features onto the elevation BEV to include appearance information. Radar-to-LiDAR \cite{yin2021radar} applies a joint learning strategy to acquire robust features that can be used in cross-modality localization. Other than BEV representation, 3D volume is also an effective spatial representation. In \cite{harley2022simple}, 3D feature volume is used to fuse multi-modal information, but the high demand for computational memory prevents its application. In this study, we introduce BEV representation to tackle the VPR task. With all features represented in a unified BEV representation, our method can be easily extended to multi-modal scenarios.

\begin{figure*}[tbp]
	\centering
		\includegraphics[width=\linewidth]{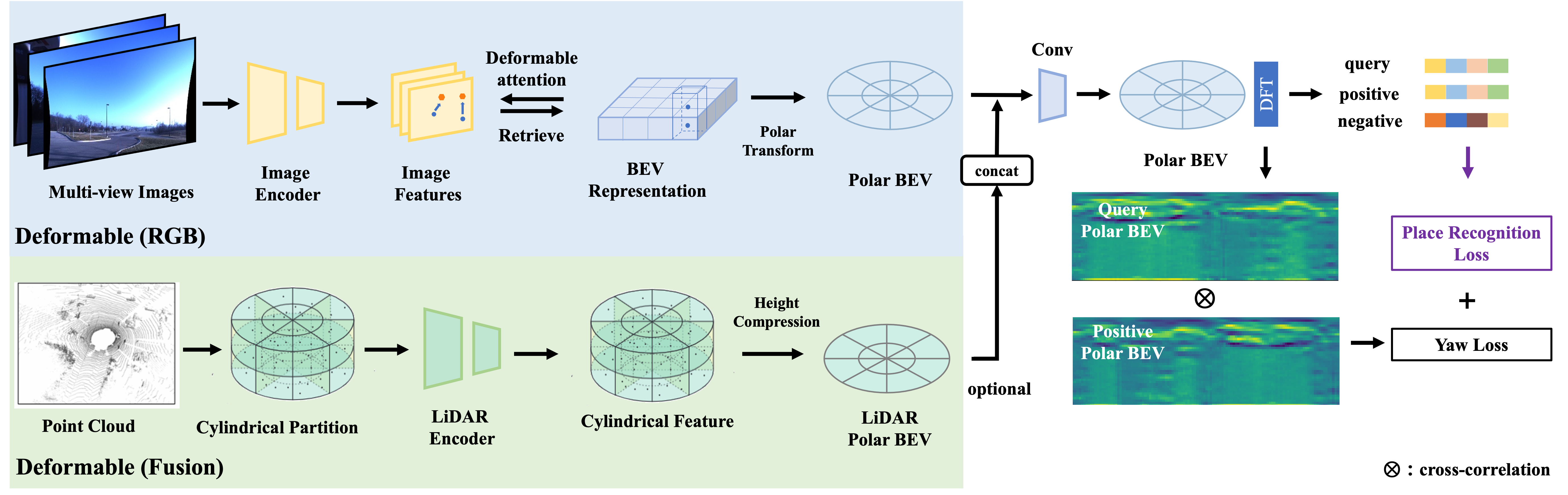}
	\caption{Overview of our framework. Given multi-view images, an image encoder is applied to acquire image features and then sampled points in the BEV representation are projected onto the image plane and associated features are gathered in a BEV. To make use of the DFT for rotation invariance, the Polar transform is leveraged to convert rotation to translation dimension.}
	\label{fig:pipeline}
    \vspace{-0.2cm}
\end{figure*}

\section{Overview}

Our BEV-based 360-degree visual place recognition network takes observations of a place $p$, including multi-camera images, denoted as $I_p = \{I^{i}_p\}^{N_{view}}_{i=1}$, where $I^{i}_p$ is the image of the $i$-th view, $N_{view}$ is the total number of camera views. After feature extraction, the multi-image cues are combined into a BEV representation, which is then aggregated to generate the place feature $\mathcal{M}_p$. The architecture is demonstrated in Fig.~\ref{fig:pipeline}.

Following this way, we generate the place features for all places in the existing map to build a database $\{\mathcal{M}_i\}$. We also generate the place feature for the current query place, say $\mathcal{M}_p$. By comparing the Euclidean distance between the query place feature and the features in the database, we finally recognize the current place as the one in the map having the minimal difference, say the $i^*$th place as:
\begin{equation}
    i^* = \arg\min_i \|\mathcal{M}_p - \mathcal{M}_i\|
    \label{pr}
\end{equation}

When a LiDAR measurement is available simultaneously, we further utilize a point cloud feature extractor to encode the structural feature and project the feature into BEV representation. As both modalities are represented in the same spatial coordinates, we simply concatenate the BEV representations for place feature aggregation. We assume that the sensory data is synchronized and that the intrinsic and extrinsic parameters are known.

\section{BEV Representation Encoder}
\label{method}

The type of camera used in multi-camera platforms may differ since each camera serves a different role. For example, in the Oxford Radar Dataset \cite{barnes2020oxford}, the city autonomous driving scenario demands that the front view is significantly more important than other views, therefore, the front view is covered by a high-resolution stereo camera, whereas the three side-view cameras are fisheye. With the known intrinsic parameters, the fisheye images can be easily undistorted. As for the front stereo camera, the short baseline resulted in a huge overlap of view, therefore, we only utilize one of the stereo images to cover the area in the front. All images are cropped and resized to fit the GPU memory.

\subsection{Image Feature Extraction}

After the image processing step, we then extract features $F_p = \{F^{i}_p\}^{N_{view}}_{i=1}$ from each one of the processed multi-camera images $I_p$. The visual feature could be extracted by any commonly used 2D backbone, such as ResNet \cite{he2016deep} or EfficientNet \cite{tan2019efficientnet}. Note that we do not stitch the images into one panorama but keep images undistorted separately, the reason is that such image formation is the best match for existing standard convolutional network-based feature extraction backbones, while for panorama, special architecture is required, preventing the utilization of pre-trained model.

\subsection{Vanilla BEV Representation}
\label{geo_repre}

To integrate the multi-image features into a unified feature, we employ BEV representation, which endows the feature with spatial awareness. Compared with the concatenation of multi-image features, we consider that such spatial awareness provides a unified frame for multi-images fusion, bringing a stronger inductive bias on geometric constraints. 

We define a 3D volume representation $\mathcal{G}$ fixed to the center of the multi-camera system to gather all image features. Its horizontal plane is aligned with the BEV plane.
After the image feature extraction, each center $g_j$ of these voxel grids in $\mathcal{G}$ is projected to the image plane using the known camera parameters $\mathcal{K}$ and retrieves the image features. To avoid quantization error, bi-linear sampling is utilized to handle subpixel retrieval. We denote the views having a valid point projection as $\mathcal{V}_{p}$ i.e. point lies inside the camera view frustum. The process steps can be formulated as:
\begin{equation}
\label{bev}
\mathcal{G}(g_j) = \frac{1}{|\mathcal{V}_p|}\sum_{k\in \mathcal{V}_p}{F^{k}_{p}{(\mathcal{P}(g_j,k,\mathcal{K}))}}
\end{equation}
where $k$ indexes the projected camera view, $F^k_p$ is the feature map of the $k$-th camera image. We use camera parameters $\mathcal{K}$ to form a projection function $\mathcal{P}(g_j,k, \mathcal{K})$ to get the image position of a voxel grid on the $k$-th image. For the voxel having multiple features retrieved from different camera views, we average these features. The averaged 3D features are then compressed to a BEV using convolution layers along the height dimension. We use this BEV feature representation as our vanilla version.

Since the retrieval step requires accurate camera parameters, inaccurate calibration may have negative effects on the final results. Synchronization error is another negative source for the vanilla version feature construction. To alleviate these negative effects, we adopt deformable attention, which is an efficient attention layer where each query can interact with different parts of the image by a learned offset compensating the misalignment.

\begin{figure}[tbp]
	\centering
		\includegraphics[width=7cm]{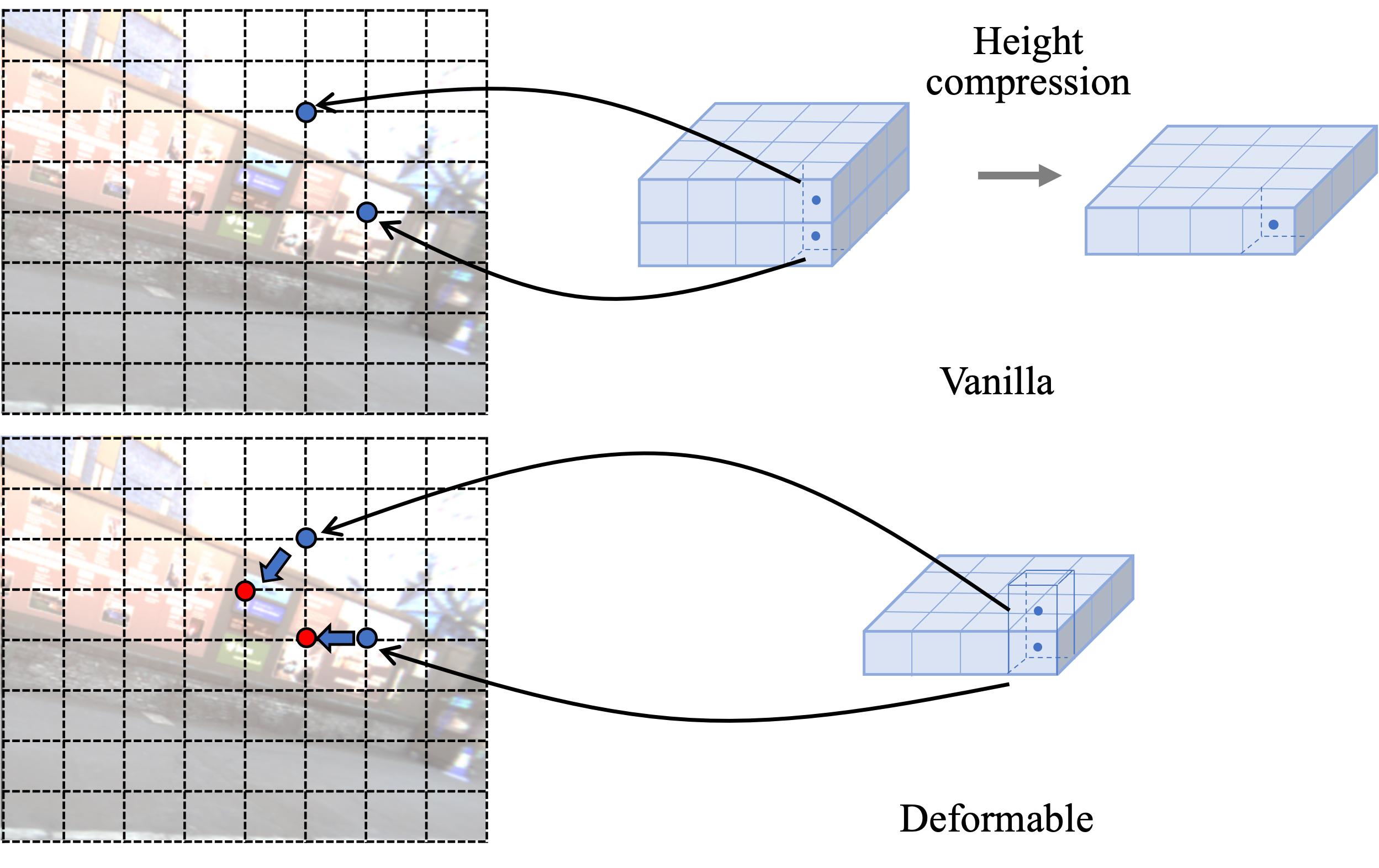}
	\caption{Demonstration of the vanilla and deformable attention. In the vanilla version, the centers of voxel grids are projected onto fixed image pixels as blue dots. In the deformable version, the sampled points in BEV are first projected onto the image shown as blue dots, and then the learnable offsets (blue arrows) finetune the position of the pixels to the red dots.}
	\label{fig:deformable}
    \vspace{-0.2cm}
\end{figure}

\subsection{Deformable Attention for BEV Representation}
We present the spatial cross-attention based on deformable attention to address the misalignment as well as the high computational cost imposed by the large input scale of multi-camera images. As shown in Fig.~\ref{fig:deformable}, instead of a fixed projected pixel, the deformable attention allows each voxel grid to project on different areas, which can fall on some views determined by the camera parameters and learned offsets. Specifically, the deformable attention mechanism is formulated as:
\begin{equation}
\begin{aligned}
\begin{split}
\label{deformattn}
\Psi(q_j, p, F) =\sum^{N_{head}}_{i=1}W_i\sum^{N_{key}}_{j=1}A_{ij}\cdot W^{\prime}_iF(p+\Delta p_{ij})
\end{split}
\end{aligned}
\end{equation}
where $q_j$, $p$, $F$ represent the query, sampled pixel and feature, respectively. Precisely, the query is a learnable parameter and sampled pixel is the corresponding pixel of the query. $N_{head}$ is the total number of attention heads and $N_{key}$ is the total number of sampled keys for each head. $W_i$ and $W^{\prime}_i$ are the learnable weights. $A_{ij}$ is the normalized attention weight and $\Delta p_{ij}$ is the predicted offsets for this sampled point. $F(p+\Delta p_{ij})$ is the retrieved feature at positions $p+\Delta p_{ij}$. Note that $\Delta p_{ij}$ allows for the compensation of the calibration and synchronization error.

To avoid large memory consumption of 3D volume representation $\mathcal{G}$, we eliminate the height dimension of the $\mathcal{G}$ to get a lightweight BEV representation, $\mathcal{B}$. Same as the definition in \cite{li2022bevformer}, in the BEV representation, we first predefine a 2D BEV grids and a group of learnable parameters $Q=\{q_j\}$ as the queries, and each query is located at each grid $b_j$ of the BEV. To avoid information loss in the height dimension, each query on the BEV plane has $N_{h}$ sampled 3D points. These sampled points act the same as the grid centers in $\mathcal{G}$. As we sum the features retrieved by $N_h$ sampled points and discard the middle 3D representation, we save a significant amount of memory. The feature retrieval process is formulated as:
\begin{equation}
\begin{aligned}
\begin{split}
\label{deformattn_bev}
\mathcal{B}(q_j, b_j) =
\frac{1}{|\mathcal{V}_p|}\sum_{k\in \mathcal{V}_p}\sum^{N_h}_{h=1}{\Psi(q_j,\mathcal{P}(b_j,h,k,\mathcal{K}), F^k_p)}
\end{split}
\end{aligned}
\end{equation}
where $k$ indexes the projected camera view and $h$ indexes the sampled point in the height dimension. For each BEV query $q_j$, we use the projection function $\mathcal{P}(b_j,h,k,\mathcal{K})$ to get the $h$-th sampled points on the $k$-th camera view. The possible misalignment in $\mathcal{P}(b_j,h,k,\mathcal{K})$ can be alleviated by the learned offset $\Delta p_{ij}$ in the (\ref{deformattn}).

\section{Aggregation and Fusion}

Aggregation aims at building the place-level feature. At one place, two visits may happen at different times and vary in perspective. The first factor, visits at different times result in appearance change which is mainly dealt with in the section on feature learning above. For the second factor, perspective change, we introduce our aggregation method, which keeps the place-level feature invariant to perspective change without learning. Moreover, the aggregation method is differentiable, enabling the back-propagation to the upstream feature learning, and the perspective change will not influence the learning of appearance features.

\subsection{Rotation Invariance}

Note that perspective change is the rotation in BEV representation, we thus apply the discrete Fourier transform (DFT) $\mathcal{F}_t$ to the polar BEV representation to achieve rotation invariance \cite{xu2021disco}. Specifically, the rotation invariance is realized by the translation invariant property of the magnitude spectrum on polar BEV, where the translation indicates the rotation in the original BEV. Denote the polarized BEV image as $\mathcal{B}_\rho(\theta, r)$, where $\theta$ is the rotation of the original BEV $\mathcal{B}$, $r$ is the range along the ray $\theta$, the invariance property can be formulated as:
\begin{equation}
\begin{aligned}
\begin{split}
\label{invariant_property_t}
|\mathcal{F}_t(\mathcal{B}_\rho(\theta,r))|=|\mathcal{F}_t(\mathcal{B}_\rho(\theta-\alpha,r))|
\end{split}
\end{aligned}
\end{equation}
where $\alpha$ is an arbitrary rotation perturbation, $|\cdot|$ is the magnitude operation. 

\begin{figure}[tbp]
	\centering
		\includegraphics[width=9cm]{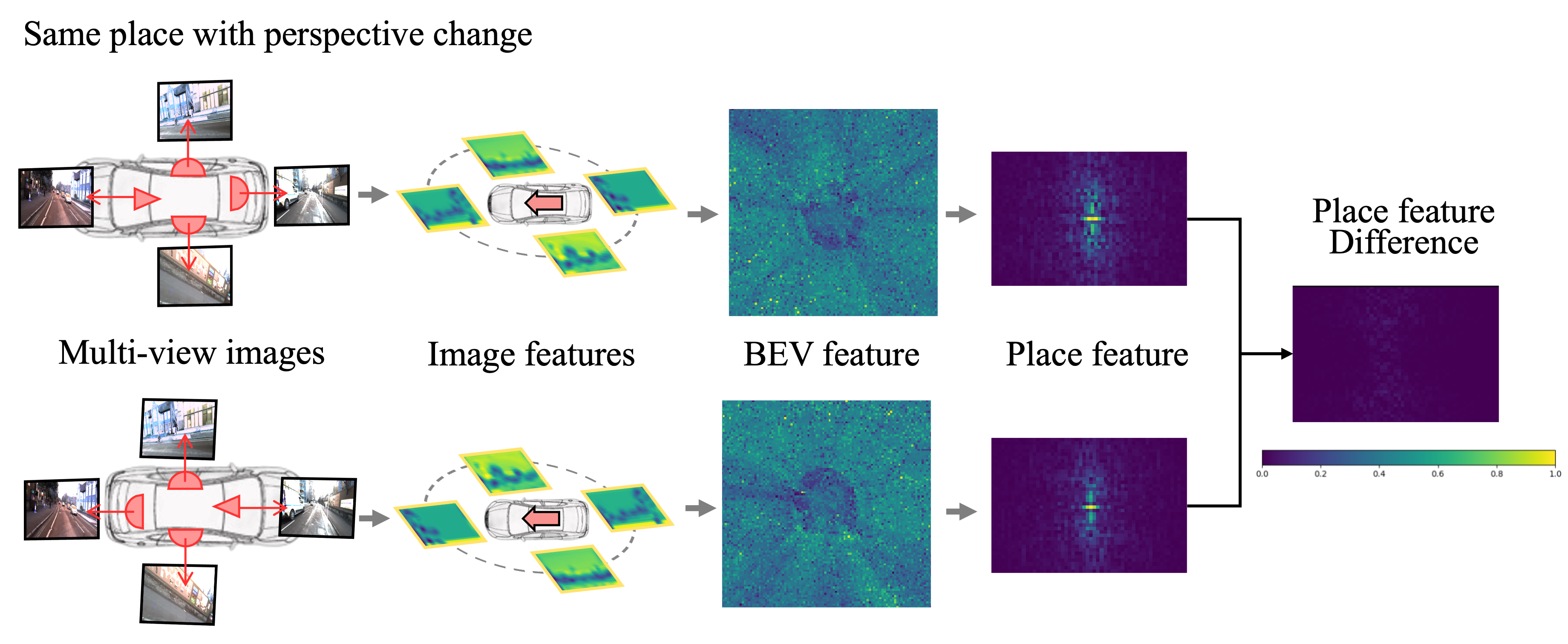}
	\caption{Case study for rotation invariance. In this case, multi-images were captured with large perspective changes (about 180-degree). The resultant place feature difference is almost zero. }
	\label{fig:toycase}
    \vspace{-0.2cm}
\end{figure}

In our method, image features are extracted on the image plane and projected to the BEV plane, forming a BEV feature representation. To improve efficiency, we first use convolution layers to reduce the BEV features' channel. With the single-layer BEV features acquired, we apply the polar transform, the DFT and magnitude, finally arriving at the place feature $\mathcal{M}$:
\begin{equation}
\begin{aligned}
\begin{split}
\label{aggregation}
\mathcal{M} = |\mathcal{F}_t(\mathcal{B}_\rho(\theta,r))|
\end{split}
\end{aligned}
\end{equation}
According to (\ref{invariant_property_t}), we know that if two visits to the same place have different perspectives but with no other changes, their place features $\mathcal{M}$s are the same. A case study is shown in Fig.~\ref{fig:toycase}. For the other changes, say environment change, measurement noise and occlusion, we leave them to the invariant feature learning.

When comparing the distance between two place features, we only use the centering $16\times16$ region of $\mathcal{M}$ for efficiency, because we find that the majority of useful data exists in the low frequency of the magnitude of the frequency spectrum. In addition, note that the Euclidean distance-based search can be implemented by the KD-tree, and the efficiency of database-level comparison, i.e. place recognition (\ref{pr}), becomes tractable.

\subsection{Yaw Estimation}

Different from the methods that only generate global features, in our BEV representation, it is also possible to estimate the perturbation $\alpha$, according to (\ref{invariant_property_t}). Given the current place observation and the retrieved features, the perturbation is actually the relative yaw angle, i.e. perspective change, between the two visits. 

Formally, given the two polar BEV visiting the same place with only perspective change, $\mathcal{B}_{p,\rho}$ and $\mathcal{B}_{i^*,\rho}$, we estimate yaw angle by the cross-correlation $\otimes$ following \cite{xu2021disco,lu2022one, xu2022ring++}:
\begin{equation}
\begin{aligned}
\begin{split}
\label{cc}
    S =&\mathcal{B}_{p,\rho}(\theta,r) \otimes \mathcal{B}_{i^*,\rho}(\theta,r)\\
    = &\mathcal{B}_{p,\rho}(\theta,r) \otimes \mathcal{B}_{p,\rho}(\theta-\alpha,r)
\end{split}
\end{aligned}
\end{equation}
As the rotation of BEV is the translation in polar BEV, the translation of the peak in $S$ tells the relative yaw angle $\alpha$. To accelerate the computation of cross-correlation, we apply DFT on both sides of (\ref{cc}):
\begin{equation}
    \mathcal{F}_t(S) = \mathcal{F}_t(\mathcal{B}_{p,\rho}) \odot \mathcal{F}_t(\mathcal{B}_{i^*,\rho})
\end{equation}
where $\odot$ is the element-wise multiplication. By extracting the phase of $\mathcal{F}_t(S)$, denotes $\angle \mathcal{F}_t(S)$, and applying the inverse DFT, we should have a Dirac function peaking at $\alpha$. This process is called phase correlation. In practice, the change between two visits cannot be perfect as an assumption, so we apply a $softmax$ layer to build a distribution of $\alpha$:
\begin{equation}
    p(\alpha) = softmax(W \cdot \mathcal{F}_t^{-1}(\angle \mathcal{F}_t(S)) + b)
\end{equation}
where $W$ and $b$ are two temperature parameters for $softmax$. Unlike the $\max$ operation, another advantage of $softmax$ is that it is differentiable, allowing for the back-propagation from loss to upstream feature learning.




\subsection{Vision-LiDAR Fusion}
BEV representation is popular in recent LiDAR-based place recognition methods \cite{kim2018scan,xu2021disco,xu2022ring++} and achieves superior performance. When LiDAR is available, thanks to the BEV feature representation, we can simply concatenate the LiDAR feature as different modalities share the same spatial coordinates. This can be an additional advantage of BEV representation compared with the place feature using vector-level fusion \cite{komorowski2021minkloc++,lai2021adafusion}. 

Specifically, we first transform the input LiDAR points into a polar coordinate and then adopt a sparse convolutional network to acquire cylindrical features. A height compression module is further applied to form a polar BEV that is consistent with the visual BEV. Then, we concatenate the features in each grid of the polar BEV and get the multi-modal BEV. The architecture of the vision-LiDAR fusion pipeline is shown in Fig.~\ref{fig:pipeline}. With the unified BEV representation, the aggregation part of the vision-LiDAR fusion remains the same. We verify the advantage of fusion in the same coordinates in the experiments.

\section{Network Training}

We crop the original image and only keep the center part because the images from datasets contain a large portion of the sky (especially in the NCLT dataset) that is useless for the VPR task. The image is then resized to $224\times384$ for the NCLT dataset. As for the Oxford dataset, we use the center view of the front stereo camera and three other view images. These images are undistort and resized to $320\times640$. Note that, as the image is cropped, the given camera parameters should be modified correspondingly.

\begin{figure}[tbp]
	\centering
		\includegraphics[width=9cm]{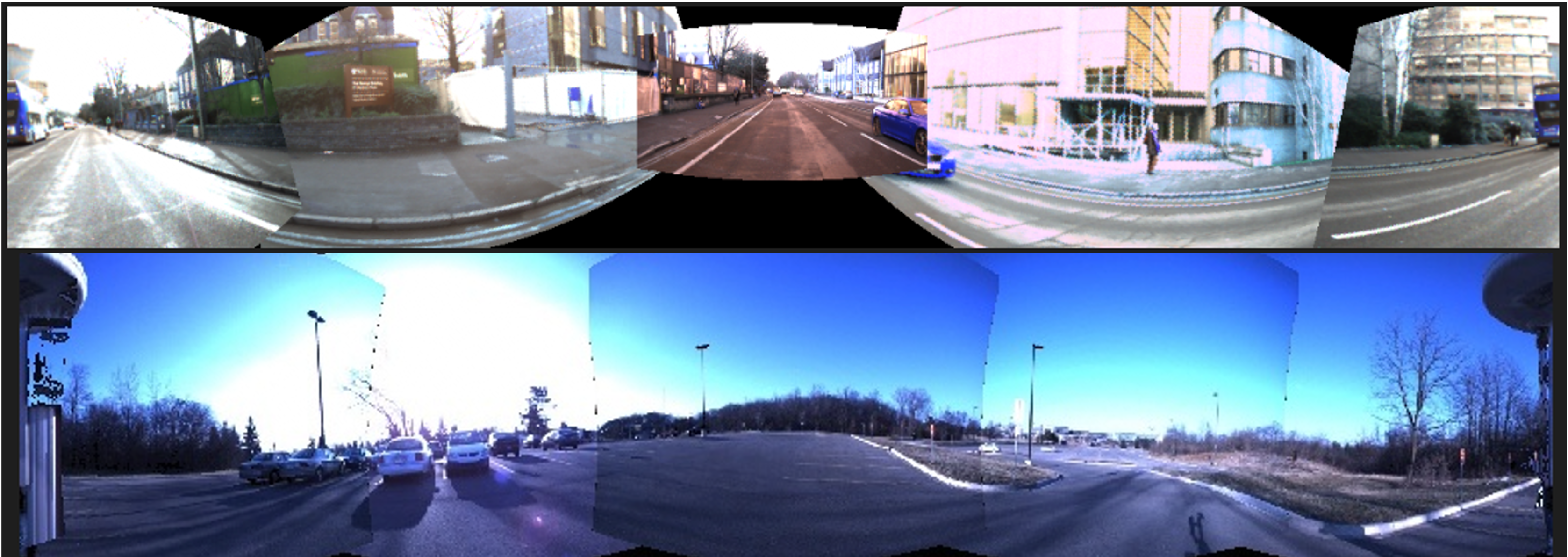}
	\caption{Generated panoramic images from the Oxford Radar dataset (top) and the NCLT dataset (bottom).}
	\label{fig:panoramas}
    \vspace{-0.2cm}
\end{figure}

Panoramic images are generated using sampling methods. We first define spherical grids and then project those grids onto the multi-view images using camera parameters to acquire the corresponding pixel value. Two generated panoramic images are shown in Fig. \ref{fig:panoramas}. Due to the inaccurate calibration results, the panoramic images generated from the Oxford Radar dataset suffer from misalignment in the seams between images. The size of the panoramic image used in comparative methods is $160\times768$ for the NCLT dataset and $200\times1280$ for the Oxford dataset.


\subsection{Loss and Training Settings}
To train the network for discriminative features learning, we follow the common practice \cite{komorowski2021minkloc++, komorowski2021minkloc3d, xu2021disco} to adopt metric learning with triplet margin loss. Multi-view images and optionally a corresponding 3D point cloud form a mini-batch. Each batch consists of several mini-batches that can be divided into an anchor, positive and negative examples. Positive examples are those that are within $2m$ of the anchor, whereas negative examples are at least $3m$ apart. We use batch-hard negative mining to eliminate triplets with zero losses in order to improve training efficiency. The loss term is given as:
\begin{equation}
\begin{aligned}
\begin{split}
\label{prloss}
\mathcal{L}_P(s_i,s_i^+,s_i^-) = max\{||s_i - s_i^+||_2-||s_i - s_i^-||_2 + m, 0\}
\end{split}
\end{aligned}
\end{equation}
where $s_i$, $s_i^+$, $s_i^-$ are place features of the anchor, positive and negative examples in the $i$-th batch. $m$ is the margin of the triplet loss, which is set to $0.2$ in our experiments.

As for yaw estimation, we use the loss term as KL divergence between the probability distribution $p(\alpha)$ and the supervision of a one-hot distribution $\textbf{1}(\alpha^*)$ peaking at the ground truth $\alpha^*$.
\begin{equation}
\begin{aligned}
\begin{split}
\label{yawloss}
\mathcal{L}_{yaw} = KLD(p(\alpha),\textbf{1}(\alpha^*))
\end{split}
\end{aligned}
\end{equation}

We train our network in a joint manner. Two losses (\ref{prloss}) and (\ref{yawloss}) are combined into a joint loss $\mathcal{L}$:
\begin{equation}
\begin{aligned}
\begin{split}
\label{totalloss}
\mathcal{L} = \mathcal{L}_P + \lambda \mathcal{L}_{yaw}
\end{split}
\end{aligned}
\end{equation}
In our experiment, the $\lambda$ is set to $0.001$. In all learning-based experiments, we train the network for 30 epochs, reducing the learning rate by 10 at the end of 20 epochs. In the VPR task, the dimension of the image feature is set to 256, while in the multi-modal experiments, the dimensions of the image and point cloud features are set to 128.

\section{Experiments}

\subsection{Dataset and Evaluation}

We apply our method to multi-session datasets with multi-modal information, the NCLT, and the Oxford Radar RobotCar datasets. Both datasets provide multi-view images to offer 360-degree VPR. Furthermore, we utilize different sequences for multi-session place recognition evaluation. The characteristics of these sequences are introduced in the subsections. 

\subsubsection{NCLT Dataset \cite{carlevaris2016university}}

A large-scale and long-term dataset collected on the University of Michigan's North Campus by a Segway robot. It includes 27 separate sessions that were recorded biweekly between January 8, 2012, and April 5, 2013. The dataset, which spans 15 months and includes a wide range of environmental changes, includes dynamic items such as moving individuals, seasonal changes such as winter and summer, and structural changes such as building construction. We choose ``2012-02-04'' and ``2012-03-17'' sessions for training and testing, with ``2012-02-04'' serving as the database session and ``2012-03-17'' serving as the query session. 

\subsubsection{Oxford Radar RobotCar Dataset \cite{barnes2020oxford}}
An addition to the \textit{Oxford RobotCar Dataset \cite{maddern20171}} for the study of multi-modal tasks. In January 2019, 32 traversals of a central Oxford route were recorded. This dataset includes a wide range of weather and lighting situations. To increase 3D scene understanding performance, a pair of Velodyne HDL-32E 3D LiDARs are mounted on the vehicle's left and right sides. The 360-degree vision is provided by a front stereo camera and three fisheye cameras. We concatenate the point clouds recorded by these two LiDARs into a single scan for simplicity in place recognition evaluation. We use ``2019-01-11-13-24-51'' as a database session and ``2019-01-15-13-06-37'' as a trajectory for training and testing, where ``2019-01-11-13-24-51'' serves as a database session and ``2019-01-15-13-06-37'' serves as a query session. Note that the given camera parameters and the synchronization across sensors in the dataset are not perfect. 

To avoid repeating the process in the same place when the vehicle does not move, we follow the strategy in \cite{komorowski2021egonn} which ignores consecutive scans with less than $20cm$ intervals. The characteristics of the two datasets are summarized in Tab. \ref{tab:dataset_detail}.

\begin{table}
\renewcommand\arraystretch{1.3}
\centering
\caption{Dataset details}
\label{tab:dataset_detail}
\begin{tabular}{lcccc}
\toprule

\multirow{2}{*}{Dataset} & \multirow{2}{*}{Texture} & Environmental & Rotation  & \multirow{2}{*}{Synchronization} \\ 

& & Changes & Variance & \\ \hline

NCLT & Sparse & $\star\star\star$ & $\star\star\star$ & Good \\ \hline

Oxford & Rich & $\star$ & $\star$ & Noisy \\

\bottomrule
\end{tabular}
\vspace{-0.2cm}
\end{table}

\subsubsection{Evaluation Metrics} In the experiment section, we evaluate the performance of the global features and yaw estimation. We follow the similar evaluation protocal as in \cite{arandjelovic2016netvlad, uy2018pointnetvlad}. As mentioned in the description of each dataset, the evaluation dataset is made up of a query and a database set that cover the same trajectory but are from different sessions. We use $Recall@N$ to evaluate the performance. It measures the percentage of successfully localized queries using the top $N$ candidates retrieved from the database. Localization is successful if one of the top $N$ retrieved candidates is within $d$ meters of the ground truth. In most of our experiments, $d$ is set to 2m.

\subsection{Comparative Methods}

\begin{itemize}
    \item \textbf{VPR methods:} We compare a series of learning-free VPR methods that use handcrafted local features (ORB \cite{rublee2011orb}, SIFT \cite{valgren2007sift}, DenseSIFT \cite{liu2008sift}) with ASMK \cite{tolias2013aggregate} and learning-based VPR methods NetVLAD \cite{arandjelovic2016netvlad}, DOLG \cite{yang2021dolg} and HowNet with ASMK \cite{tolias2020learning}. There are two inputs of all comparative VPR methods, panoramic and front-camera image.
    
    \item \textbf{Vision-LiDAR fusion methods:} We further evaluate our methods on the vision-LiDAR fusion scenarios to confirm the capability to fuse various sensor data. We also assess the performance of some recent vision-LiDAR fusion methods, such as Minkloc++ \cite{komorowski2021minkloc++} and AdaFusion \cite{lai2021adafusion}. 
\end{itemize}

For fast convergence, we utilize pre-trained feature extraction modules in all methods. In the NetVLAD experiments, we also finetune the model pre-trained on the Pittsburgh dataset \cite{torii2013visual}. 

In addition, we evaluate the vanilla version of our method, which adopts 3D volume as the middle representation and simply retrieves the grids' features by directly projecting the grids' locations onto the image plane, the detailed description can be found in Section \ref{geo_repre}. 

\begin{figure}[tbp]
	\centering
		\includegraphics[width=9cm]{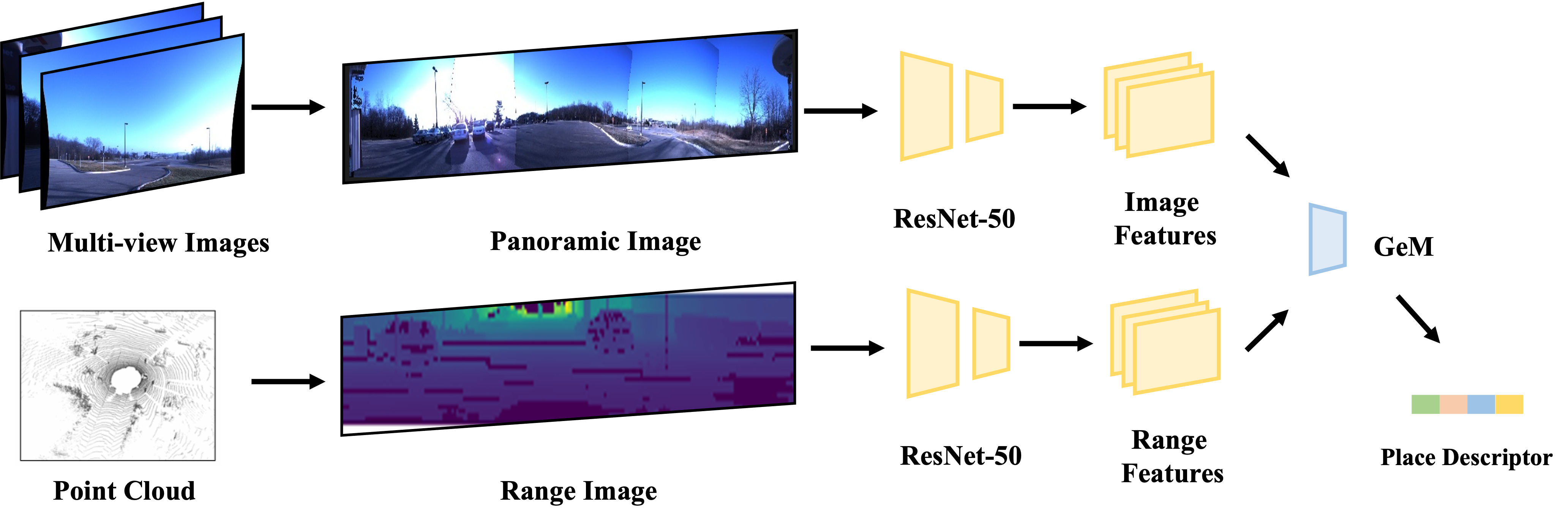}
	\caption{Demonstration of the panorama pipeline.}
	\label{fig:panorama_pipeline}
    \vspace{-0.2cm}
\end{figure}

\subsection{Verification of BEV Representation}

We hypothesize that BEV representation can play an important role in feature extraction, feature aggregation and vision-LiDAR fusion. Thus we first design the study to show the efficacy of the BEV representation in these three aspects. We build a panorama pipeline with standard CNN-based feature extraction to suppress the influence of network architecture. We use panoramic and range images as input for the sensor fusion experiment and use the same feature extraction and aggregation components. The outputs of two branches are then concatenated to create a global place feature, as in \cite{komorowski2021minkloc++}. Specifically, the feature extraction module is ResNet-50, and the aggregation method is the image level GeM pooling \cite{radenovic2018fine}. The implemented pipeline is demonstrated in Fig.~\ref{fig:panorama_pipeline}. Therefore, the image-level representation and the BEV representation can be compared.

We report the performance of our method with deformable attention and the panorama pipeline, in terms of place recognition. As shown in Tab \ref{tab:representation}, the BEV representation with $Recall@1$ of 63.6\% outperforms the panorama representation with $Recall@1$ of 58.7\%. This enhancement verifies that the spatial awareness of BEV representation can benefit the feature extraction and aggregation in 360-degree VPR. We can also find the same result in vision-LiDAR fusion experiments. As for the comparison between the vision-only and vision-LiDAR fusion, it is natural that all pipelines are improved significantly. When features are represented in BEV, both GeM pooling and DFT provide invariance, but the information loss in GeM pooling is serious. DFT, on the other hand, preserves the low-frequency information which corresponds to the primary portion of the image, thus achieving a good performance in both vision-only and fusion pipelines. 

\begin{figure*}[tbp]
	\centering
		\includegraphics[width=\linewidth]{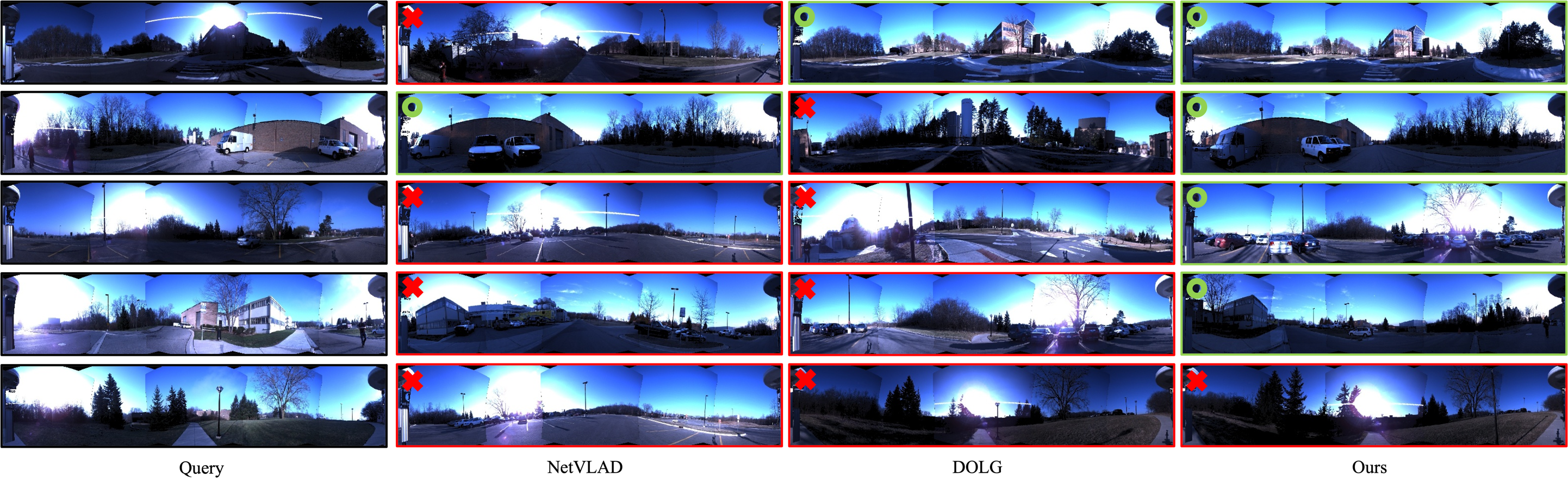}
	\caption{Examples of top-1 retrieval of different methods. The panoramic images are used for better visualization. We demonstrate some retrieval cases with significant perspective and environmental changes. Note that the revisit criterion is $2m$, so some of the false retrievals may also be similar to the query.}
	\label{fig:cases}
    \vspace{-0.2cm}
\end{figure*}

\subsection{Comparison with Baseline Methods}
We evaluate all methods using identical evaluation protocols. The evaluation results on the NCLT and the Oxford Radar RobotCar dataset are shown in Tab. \ref{tab:vpr_result}. The top two parts of the table are results from learning-free methods and the left are learning-based methods. In the Input Mode column. \textit{Front} represents Front-camera image and \textit{Pano} represents the Panoramic image. 

A quick finding is that 360-degree sensing brings improvement to the front camera image only. When we compare the performances between datasets, most methods perform better on the Oxford dataset. The Oxford dataset was collected in a city scenario with no significant environmental changes and rotation variance within a week. The rich texture benefits the feature extraction module, resulting in better overall place recognition performance. As the results demonstrate, in such simple scenarios, handcrafted methods based on the human experience are competitive with learning-based methods. Among the learning-based methods, ours performs better in $Recall@1$ and about the same in $Recall@5$. 

The NCLT dataset, on the other hand, is more challenging, as there exist significant environmental changes and contains many hard scenarios with sparse texture. As a consequence, all competing methods exhibit severely degraded performance, while ours demonstrates almost similar performance to that in the Oxford dataset. We explain this as the inductive bias brought by spatial awareness, which enforces feature learning by suppressing the impact of perspective change. Some retrieval examples are shown in Fig.~\ref{fig:cases}. The chosen scenarios include severe illumination changes, perspective changes as well as environmental changes involving seasonal change and dynamic objects. The third row, in particular, demonstrates that our method is not sensitive to dynamic objects and the fourth case shows the strong rotation invariance of our method. However, in the last scenario, there are plants everywhere, making all methods difficult. For other cases, thanks to spatial awareness, the retrievals of our method are closer to the query place than those of comparative methods. In addition, the improvement between the vanilla version and the deformable version verifies the compensation for hardware errors. Deformable attention provides a solution to the problem of inaccurate synchronization and calibration. As shown in Tab. \ref{tab:vpr_result}, in the NCLT dataset, the deformable attention helps find better corresponding features, while in the Oxford dataset, the deformable attention is used to alleviate the negative effects brought by the inaccurate calibration.


\textbf{Revisit criterion}
The revisit criterion determines whether the query is successfully recognized. To verify the robustness against the translation of place recognition, we use different revisit criteria ranging from $2m$ to $20m$. We notice that our methods outperform others across all thresholds. Our method with deformable attention has a smoother trend than others, which infers that the incorrectly retrieved places are far away from the query place. Based on this finding, we consider that our methods are insensitive to the selection of places in the map database.

\begin{table}
\renewcommand\arraystretch{1.4}
\centering
\caption{Comparative study on representation. Evaluated on the NCLT dataset}
\label{tab:representation}
\begin{tabular}{lccc}
\toprule

Representation & Aggregation & Recall@1 & Recall@5   \\

\midrule

Panorama (RGB) & GeM & 58.7 & 77.3\\ 
BEV (RGB) & GeM & 63.6 & 78.9 \\
BEV (RGB) & DFT & \textbf{81.3} & \textbf{90.2} \\ \hline

Panorama (Fusion) & GeM & 77.8 & 91.0\\ 
BEV (Fusion) & GeM & 81.5 & 94.5 \\
BEV (Fusion) & DFT & \textbf{91.6} & \textbf{97.4} \\ 

\bottomrule
\end{tabular}
\vspace{-0.4cm}
\end{table}

\begin{table*}
\renewcommand\arraystretch{1.3}
\centering
\caption{Quantitative Results of Visual Place Recognition}
\label{tab:vpr_result}
\begin{threeparttable}
\begin{tabular}{clccccccc}
\toprule
\multicolumn{1}{c}{\multirow{2}{*}{Type}} & \multicolumn{1}{l}{\multirow{2}{*}{Methods}} & \multicolumn{1}{c}{\multirow{2}{*}{Visual Input Mode}} & \multicolumn{2}{c}{NCLT} & \multicolumn{2}{c}{Oxford Radar} & \multicolumn{2}{c}{Mean}  \\ \cline{4-9}


& & &\multicolumn{1}{c}{R@1} & \multicolumn{1}{c}{R@5} & \multicolumn{1}{c}{R@1} & \multicolumn{1}{c}{R@5} & \multicolumn{1}{c}{R@1} & \multicolumn{1}{c}{R@5} \\ \hline

\multirow{6}{*}{Non-learning VPR} & ORB + ASMK  & front & 17.9 & 26.4 & 56.1 & 78.3 & 37.0 & 52.4 \\
& SIFT + ASMK & front & 29.6 & 39.6 & 78.8 & 91.0 & 54.2 & 65.3\\
& DenseSIFT + ASMK & front & 20.6 & 29.3 & 75.2 & 87.9 & 47.9 & 58.6\\ \cline{2-9}

& ORB + ASMK & pano & 8.9 & 17.4 & 67.7 & 84.6 & 38.3 & 51.0\\
& SIFT + ASMK & pano & 20.3 & 33.0 & \textbf{85.6} & \textbf{95.5} & 53.0 & 64.3\\
& DenseSIFT + ASMK & pano & 13.1 & 17.9 & \underline{84.1} & \underline{93.1} & 48.6 & 55.5 \\ \hline

\multirow{8}{*}{Learning-based VPR} & How-ASMK & front & 21.1 & 34.8 & 66.8 & 84.0 & 44.0 & 59.4\\ 
& NetVLAD$\dagger$ & front & 37.8 & 47.2 & 74.5 & 87.7 & 56.2 & 67.5\\
& DOLG & front & 20.1 & 34.3 & 53.1 & 72.8 & 36.6 & 53.6\\ \cline{2-9}

& How-ASMK & pano & 55.0 & 76.3 & 80.0 & 91.1 & 67.5 & 83.7 \\ 
& NetVLAD$\dagger$ & pano & 52.6 & 71.0 & 76.6 & 91.3 & 64.6 & 81.2 \\
& DOLG & pano & 54.5 & 75.3 & 62.0 & 82.5 & 58.3 & 78.9 \\ \cline{2-9}

& \textbf{Vanilla (Ours)} & multi-view & \underline{74.5} & \underline{84.9} & 79.0 & 88.8 & \underline{76.8} & \underline{86.9}\\ 
& \textbf{Deformable (Ours)} & multi-view & \textbf{81.3} & \textbf{90.2}  & 81.0 & 90.9 & \textbf{81.2} & \textbf{90.6}\\ 

\hline
\hline

\multirow{6}{*}{Vision-LiDAR Fusion} & Minkloc++ & front & 43.2 & 58.2 & 68.1 & 83.0 & 55.7 & 70.6\\
& AdaFusion & front & 31.0 & 46.5 & 55.7 & 75.0 & 43.4 & 60.8\\ \cline{2-9}

& Minkloc++ & pano & 72.8 & \underline{88.1} & 75.8 & 90.0 & 74.3 & 89.5 \\ 
& AdaFusion & pano & 52.2 & 71.5 & 52.8 & 75.2 & 52.5 & 73.4 \\ \cline{2-9}

& \textbf{Vanilla (Ours)} & multi-view & \textbf{92.8} & \textbf{97.4} & \underline{86.7} & \underline{95.4} & \textbf{89.8} & \underline{96.4}\\ 
& \textbf{Deformable (Ours)} & multi-view & \underline{91.6} & \textbf{97.4} & \textbf{87.2} & \textbf{96.1} & \underline{89.4} & \textbf{96.8} \\

\bottomrule
\end{tabular}
\begin{tablenotes}
    \footnotesize
    \item[$\dagger$] We finetune the NetVLAD model that was pre-trained on the Pittsburgh dataset.
\end{tablenotes}
\end{threeparttable}
\vspace{-0.4cm}
\end{table*}

\begin{figure}[tbp]
	\centering
		\includegraphics[width=9cm]{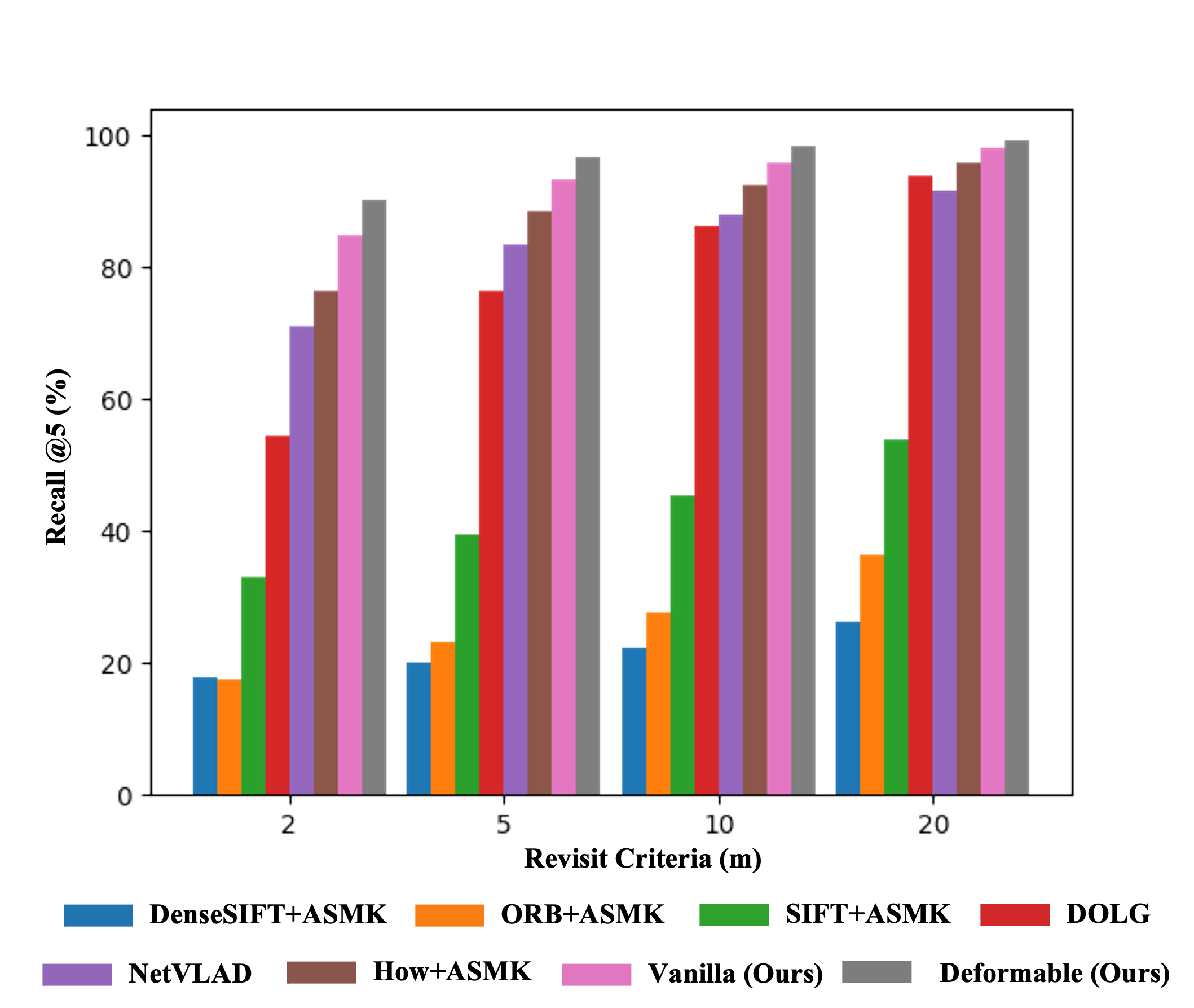}
	\caption{Recall@5 vs. Revisit Criteria on the NCLT dataset.}
	\label{fig:revisit_criteria}
    \vspace{-0.2cm}
\end{figure}

\textbf{Vision-LiDAR fusion}
The BEV representation is often used in point cloud learning. As stated in the same coordinates, we further incorporate LiDAR BEV features into our visual BEV representation and thus achieve sensor fusion. We further compare our multimodal methods with previous works. Minkloc++ \cite{komorowski2021minkloc++} and AdaFusion \cite{lai2021adafusion} were originally evaluated in the Oxford RobotCar dataset \cite{maddern20171} which similar places are defined within $25m$. We retrain and evaluate these methods with our experiment settings, where the revisit criterion is $2m$. The two modalities in Minkloc++ and AdaFusion are concatenated at the place feature level using a late fusion strategy, whereas our methods combine the modalities at the feature level using the same coordinate, which can be regarded as middle fusion. To make a fair comparison, we decrease the dimension of each modality feature to 128, forming a final multi-modal feature with the same dimension (256) of RGB-only methods. As shown in the Vision-LiDAR Fusion part of Tab. \ref{tab:vpr_result}, with the LiDAR features involved, the overall performance of our methods improves a lot. Our explanation is that we properly combine the two modalities with a unified coordinate in the feature level, which also allows for the advantage of middle fusion \cite{liu2022bevfusion}. The inconsistent coordinates at the feature level and simple late fusion, however, negatively affect other methods. In addition, referring to Tab.~\ref{tab:vpr_result}, it is a little surprising that with BEV-based extraction and aggregation, our vision-only method also achieves competitive performance with the fusion method.









\subsection{Yaw Estimation Evaluation}
As a side product, with the help of the correlation estimator, our method can provide yaw estimation between two BEVs. We assess the yaw estimation results following \cite{ding2022translation}, we demonstrate the yaw estimation errors using a quartile, which reflects the errors at the percentages of 25\%, 50\% and 75\% of
the whole results. We notice that the BEV representation not only captures the spatial relationship between features but also provides coarse estimation. With the LiDAR features utilized, the performance is further improved. Such a result is promising for compensating the yaw difference before the full degree of freedom pose estimation.

\begin{table}
\renewcommand\arraystretch{1.4}
\centering
\caption{Yaw Estimation on the NCLT dataset}
\label{tab:yaw_estimation}
\begin{tabular}{lc}
\toprule

Approach & quartile($^{\circ}$)$\downarrow$   \\

\midrule
Deformable-RGB & 1.8 / 3.9 / 6.9 \\ 
Deformable-Fusion & 1.5 / 3.3 / 6.3  \\ 

\bottomrule
\end{tabular}
\vspace{-0.4cm}
\end{table}

\subsection{Backbone Ablation}
In this section, we investigate the effects of the different backbones on our method's performance. In all experiments, unless noted, we keep the other variables the same.

\begin{figure}[tbp]
	\centering
		\includegraphics[width=\linewidth]{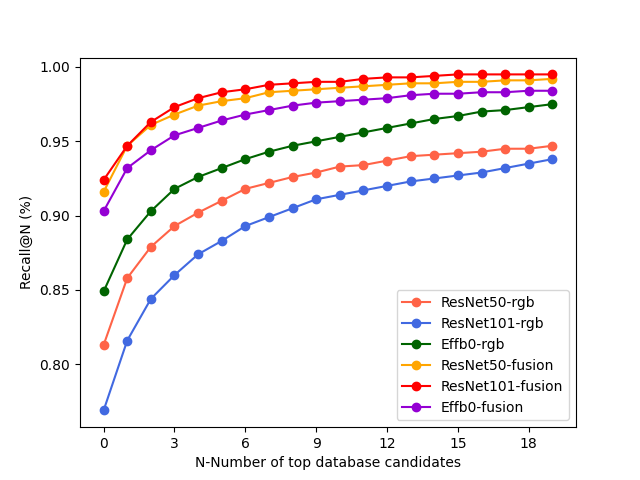}
	\caption{Recall@N(\%) of different backbones on the NCLT dataset.}
	\label{fig:backbone}
    \vspace{-0.2cm}
\end{figure}

We evaluate the performance of different image backbones. Fig. \ref{fig:backbone} shows the performance of commonly used image backbones. We find that the larger backbone does not equal better performance in RGB-only experiments. But in the fusion experiments, the larger backbone provides the better discriminative ability. We note that sometimes, the benefit of a specific backbone is tied to the input resolution which is fixed in this study. 







\subsection{Runtime Evaluation}
We assess the time cost of various modules in our approach on a GeForce RTX 2080Ti GPU with an Intel Xeon Platinum 8163 CPU. The detailed runtime analysis is listed in Tab. \ref{tab:time_cost}. The feature extraction and aggregation modules of both methods are the same. The generation time of representation is significantly reduced in the deformable pipeline due to the lightweight BEV representation and efficient spatial deformable attention. In the localization step, the place feature can be easily retrieved by using the KD-tree structure which only takes logarithm times. Therefore, our method can be incorporated into online systems.

\begin{table}
\renewcommand\arraystretch{1.3}
\centering
\caption{Computational cost of the proposed method [ms]}
\label{tab:time_cost}
\scalebox{0.9}{
\begin{tabular}{lcccc}
\toprule
Method & Image Feature & BEV Feature & Aggregation & Total \\

\midrule

Vanilla & 7.3 & 82.2 & 0.4 & 89.9 \\
Deformable & 7.3 & 48.4 & 0.4 & 56.1 \\

\bottomrule
\end{tabular}}
\vspace{-0.4cm}
\end{table}

\section{Conclusion}
In conclusion, this paper proposes the use of BEV representation for effective visual place recognition and investigates its benefits in feature extraction, feature aggregation, and vision-LiDAR feature fusion. Our proposed network architecture utilizes BEV representation to extract spatial features from images and point clouds, and achieves rotation-invariant feature aggregation through the Discrete Fourier transform. By stating image and point cloud cues in the same coordinates, our method also benefits sensor fusion for place recognition. The experimental results demonstrate that our BEV-based method outperforms baseline methods in off-the-road and on-the-road scenarios, verifying the effectiveness of BEV representation in VPR task. Therefore, our contributions in this paper demonstrate the potential of BEV representation in VPR task which can be easily integrated into the current autonomous driving framework.



\vfill

\end{document}